%% file: main.tex
\documentclass[conference]{IEEEtran}
\IEEEoverridecommandlockouts

\usepackage{cite}
\usepackage{microtype}
\usepackage{graphicx}
\usepackage{subcaption}
\usepackage{booktabs}
\usepackage{amsmath,amssymb,amsfonts}
\usepackage{mathtools}
\usepackage{amsthm}
\usepackage{algorithm}
\usepackage{algorithmic}
\usepackage{multirow}
\usepackage{multicol}
\usepackage{makecell}
\usepackage{siunitx}
\usepackage{xspace}
\usepackage[most]{tcolorbox}
\usepackage{tabularx,colortbl,xcolor}
\usepackage{listings}
\usepackage{pifont}
\usepackage{enumitem}
\usepackage[hidelinks]{hyperref}

\lstdefinestyle{pythonstyle}{
    language=Python,
    basicstyle=\ttfamily\footnotesize,
    keywordstyle=\color{blue}\bfseries,
    stringstyle=\color{red!70!black},
    commentstyle=\color{green!50!black}\itshape,
    numbers=left,
    numberstyle=\tiny\color{gray},
    numbersep=5pt,
    breaklines=true,
    showstringspaces=false,
    frame=none,
    xleftmargin=10pt,
    tabsize=4,
}

\theoremstyle{plain}

\theoremstyle{definition}

\theoremstyle{remark}

\newcommand{\methodfull}{Optimization Modeling from Solver Outcomes\xspace}
\newcommand{\methodshort}{\texttt{OMSO}\xspace}
\newcommand{\gurobi}{\texttt{Gurobi}\xspace}
\newcommand{\copt}{\texttt{COPT}\xspace}
\newcommand{\ortools}{\texttt{OR-Tools}\xspace}

\begin{document}
\bstctlcite{IEEEexample:BSTcontrol}

\title{Learning to Model Optimization Problems from Solver Outcomes}

\author{

\IEEEauthorblockN{Runda Guan}
\IEEEauthorblockA{\textit{School of Artificial Intelligence}\\
\textit{Nanjing University of Science and Technology}\\
Nanjing, China\\
rundaguan@njust.edu.cn}
\and
\IEEEauthorblockN{Xiangqing Shen$^*$}
\IEEEauthorblockA{\textit{School of Intelligence Science and Technology}\\
\textit{Nanjing University}\\
Suzhou, China\\
xqshen@nju.edu.cn}
\and
\IEEEauthorblockN{Jiajun Zhang}
\IEEEauthorblockA{\textit{Institute of Automation}\\
\textit{Chinese Academy of Sciences}\\
Beijing, China\\
jjzhang@nlpr.ia.ac.cn
}
\and
\IEEEauthorblockN{Yifan Zhang}
\IEEEauthorblockA{\textit{Institute of Automation}\\
\textit{Chinese Academy of Sciences}\\
Beijing, China\\
yfzhang@nlpr.ia.ac.cn}
\and
\IEEEauthorblockN{Jian Cheng}
\IEEEauthorblockA{\textit{Institute of Automation}\\
\textit{Chinese Academy of Sciences}\\
Beijing, China\\
jcheng@nlpr.ia.ac.cn}
\and
\IEEEauthorblockN{Rui Xia$^*$}
\IEEEauthorblockA{\textit{School of Intelligence Science and Technology}\\
\textit{Nanjing University}\\
Suzhou, China\\
rxia@nju.edu.cn}
\thanks{* Corresponding authors: Xiangqing Shen and Rui Xia.}
}

\maketitle

\begin{abstract}
Automating optimization modeling with LLMs is a promising path toward scalable decision intelligence, but existing approaches either rely on agentic pipelines built on closed-source LLMs with high inference latency, or fine-tune smaller LLMs using costly process supervision that often overfits to a single solver API.
We study how reinforcement learning from inexpensive outcome labels can improve optimization modeling and support reuse across heterogeneous solver environments.
To this end, we propose \methodfull\ (\methodshort), which uses reinforcement learning (RL) with a mathematical programming solver as a deterministic, interactive environment.
Given a natural-language problem and a target solver, \methodshort\ generates solver-specific code, executes it in a sandboxed harness, and converts execution outcomes into scalar rewards in a closed-loop \textit{generate--execute--feedback--update} process. 
This RL formulation learns optimization modeling without process-level supervision and supports cross-solver generalization by switching the solver environment rather than reconstructing solver-specific datasets.
Experiments on NL4OPT, MAMO, IndustryOR, and OptiBench across \gurobi, \ortools, and \copt\ show that \methodshort\ matches or outperforms process-supervised SFT, supports zero-shot training-free transfer, direct RL training, and SFT-cold-started RL, and improves modeling reasoning and semantic correctness beyond solver API usage.

\end{abstract}


\section{Introduction}
Data mining has long studied how to extract structured representations from unstructured or semi-structured information to support downstream computational tasks.
A major challenge arises in optimization modeling, where decision problems are often specified in unstructured natural language, while their execution relies on operations-research optimization models used in supply chains, energy systems, and finance~\cite{eskandarpour2015sustainable,skolfield2022operations,markowitz1952portfolio,SINGH2012167,doi:10.1137/1.9781611972238}.
This creates a challenging modeling step: extracting decision variables, objectives, and constraints from a \emph{natural language} problem description, constructing a valid mathematical program, and implementing it as executable solver code~\cite{meng2020mixed,ramamonjison2022nl4opt,xiao2025survey}.
The step requires careful semantic interpretation and precise mathematical formalization under solver constraints.
As a result, reliably generating correct, solver-compatible formulations from \emph{natural language} remains challenging and often still requires human experts.

Recent advances in large language models (LLMs)~\cite{grattafiori2024llama3herdmodels} in code generation and logical reasoning have opened new avenues for automating optimization modeling, with existing work mainly falling into two paradigms: prompting-based and training-based methods.
Prompting-based methods~\cite{Ahmaditeshnizi2024optimus,xiao2024chainofexperts} use multi-turn interaction, reflection, and self-correction to improve accuracy and executability, but their strongest results often rely on highly capable foundation models and multi-round inference, which can increase deployment cost and latency.
Their effectiveness may also be limited for uncommon or under-covered solvers whose APIs are weakly represented in pretraining.
Training-based methods instead fine-tune smaller LLMs for efficient inference, but reliable performance often requires fine-grained \emph{process supervision}, such as variable definitions, derivations, and reference code~\cite{huang2025orlm, wu-etal-2025-training,lu2025optmath}.
Such annotations are expensive to collect and bind the policy to solver-specific traces; when the solver interface changes, adaptation typically requires reconstructing another annotated dataset.
Thus, prompting cannot turn solver-specific failures into persistent
model improvement, whereas process-supervised fine-tuning remains costly
and brittle across solver environments.

\input{figures/framework}

To address these limitations, we propose \methodfull (\methodshort), an RL framework that treats a mathematical programming solver as a deterministic environment for optimization modeling.
Given a problem description and a target solver, \methodshort generates solver-specific executable code, runs it in a sandboxed environment, and converts the resulting solver outcome into a scalar reward.
Because solver execution exposes signals about format validity, compilation success, API compatibility, solution feasibility, and final-answer correctness, training can follow a simple \emph{generate--execute--feedback--update} loop (Figure~\ref{fig:framework}) using only problem--answer pairs.
This feedback is end-to-end: a high reward requires the generated variables, objective, and constraints to jointly define a model that can be executed, solved, and verified against the target answer.
As a result, the model can learn from inexpensive outcome supervision while still receiving task-level signals about whether its mathematical formulation is correct.
Thus, \methodshort bypasses intermediate-step traces and reference programs, replacing costly process-level modeling annotations with answer-level supervision plus solver feedback.

The same formulation also supports generalization across solver environments without rebuilding full solver-specific reference datasets.
This matters in practice because deployments may need to switch solvers due to licensing, performance, supported problem classes, or system-integration constraints.
We study three practical settings.
(i) \emph{Zero-shot training-free transfer} directly switches from a source environment to an unseen target environment without parameter updates.
(ii) \emph{Direct RL training} learns in the target environment from solver outcomes without target-solver process annotations.
(iii) \emph{SFT-cold-started RL} first uses a small translated SFT seed to establish basic API syntax for an under-covered solver, then continues RL in the target environment.
These settings cover deployment scenarios ranging from training-free solver switching to low-cost learning for solvers with limited pretraining coverage.

We conduct systematic experiments on NL4OPT~\cite{ramamonjison2022nl4opt}, MAMO~\cite{huang2025mamo}, IndustryOR~\cite{huang2025orlm}, and OptiBench~\cite{yang2025optibench}.
Across three representative solvers---\gurobi, \ortools, and \copt---we evaluate whether solver-outcome RL supports accurate optimization modeling that generalizes across solver environments.
Despite using no process annotations, \methodshort slightly surpasses the reproduced process-supervised ORLM baseline on average under the same backbone and evaluation setup.
It also supports zero-shot training-free transfer, direct RL training in different solver environments, and SFT-cold-started RL for the under-covered \copt solver.
Further ablations and human analyses suggest that the gains are associated with better modeling reasoning and fewer semantic modeling errors, rather than only improved output formatting.

\section{Related Work}

In recent years, leveraging large language models (LLMs) to automate optimization modeling has become an active research direction~\cite{xiao2025survey}.
Early progress in this line of work was driven by NL4OPT competition~\cite{ramamonjison2022nl4opt,ramamonjison2022augmenting}.
A series of benchmarks have since been introduced, including OptiBench~\cite{yang2025optibench} for diverse optimization problem types, MAMO~\cite{huang2025mamo} for evaluating mathematical modeling ability, and IndustryOR~\cite{huang2025orlm} for bringing in industrially grounded scenarios.

Existing methods largely fall into two paradigms.
Prompting-based approaches~\cite{Ahmaditeshnizi2024optimus, xiao2024chainofexperts, liu2025optitree,shi2025generalizableheuristicgenerationlarge,wei2022chain,madaan2023selfrefine,shinn2023reflexion,ye2024reevo,romera-paredes_mathematical_2024} build agentic systems through multi-step prompting, reflection, or self-correction.
While effective, these pipelines can require multiple LLM calls and therefore incur high inference cost.
Supervised fine-tuning (SFT) approaches instead train smaller open-source LLMs with intermediate-step traces that include variable definitions, formula derivations, and reference code.
ORLM~\cite{huang2025orlm} proposed the OR-Instruct data synthesis pipeline, and subsequent works~\cite{wu-etal-2025-training, lu2025optmath, jiang2025llmopt, astorga2025autoformulation, zhou2025steporlmselfevolvingframeworkgenerative} further improve dataset construction and training recipes.
However, these methods face two major challenges:
the high cost of process supervision and the difficulty of cross-solver generalization.

RL with verifiable rewards (RLVR) has recently proven effective in tasks with objective feedback, allowing training without human preference labels or learned reward models.
DeepSeek-R1~\cite{Guo_2025,shao2024deepseekmathpushinglimitsmathematical} shows that such verifiable signals can elicit stronger multi-step reasoning, and follow-up work~\cite{wen2025reinforcementlearningverifiablerewards,jin2025searchr} further supports that verifiability implicitly promotes correct reasoning behaviors.

Recent efforts have begun to explore reinforcement learning for optimization modeling.
Solver-Informed RL~\cite{chen2025solverinformed} introduces a solver-feedback-driven data synthesis pipeline to construct high-quality training samples.
OR-R1~\cite{DBLP:journals/corr/abs-2511-09092} studies test-time group-relative policy optimization, improving formulation quality via inference-time policy updates and achieving strong results with less synthesized data.
In contrast, our work studies optimization modeling from three complementary perspectives.
First, we show that modeling competence can be learned from solver outcomes alone without costly process-level traces or reference programs.
Second, we systematically study whether this competence generalizes across solver interfaces through zero-shot training-free transfer, direct RL training, and SFT-cold-started RL.
Third, we analyze whether outcome-supervised RL improves semantic modeling competence rather than merely solver API usage.
To the best of our knowledge, this is the first work to investigate cross-solver generalization under outcome-supervised RL for optimization modeling.

\section{Preliminaries}
\label{sec:preliminaries}
\input{figures/preliminary_case2}

Figure~\ref{fig:preliminary_case2} illustrates the optimization modeling task studied in this paper.
The task can be viewed through four linked elements:
\begin{itemize}
    \item \textbf{Natural-language problem.}
    The input describes a real decision problem in text, including resources, requirements, preferences, and sometimes implicit restrictions.
    It does not directly provide variables, an objective, constraints, or executable code.

    \item \textbf{Modeling rationale.}
    The modeler must identify decision variables, determine what should be optimized, and translate explicit and implicit requirements into mathematical constraints.
    This step is nontrivial because small choices, such as whether a variable is continuous or integer, can change the feasible region and final answer.

    \item \textbf{Solver code.}
    The mathematical formulation must be implemented as executable code that constructs variables, objectives, and constraints, invokes the solver, and reports the final result.
    A correct high-level formulation can still fail if the code omits a constraint, uses an incorrect variable type, mishandles the solver status, or reports the wrong value.

    \item \textbf{Solver outcome.}
    Executing the code closes the loop between the textual task and the reported answer.
    The result provides observable evidence about whether the modeling rationale and solver code jointly solve the original problem.
\end{itemize}

These elements describe successive stages of the same modeling process.
The \textbf{natural-language problem} provides the semantics of the task, the \textbf{modeling rationale} organizes them into variables, objectives, and constraints, the \textbf{solver code} implements the formulation, and the \textbf{solver outcome} reports what the implemented model produces.
Accordingly, correctness depends on both the mathematical formulation and its executable realization.

In the example, the model must infer integer cart-count variables, formulate the capacity and ratio constraints, minimize the total number of carts, and express the formulation as executable solver code.
Thus, optimization modeling is both a semantic modeling task and an executable program-generation task.

\section{Optimization Modeling from Solver Outcomes}
\label{sec:method}

Given the four elements above, one straightforward way to learn optimization modeling is to reproduce the complete expert trajectory from problem understanding to executable code.
This leads naturally to supervised fine-tuning (SFT), which imitates demonstrations that map a natural-language problem to a modeling rationale, solver code, and the expected solver outcome.
In this paradigm, training requires not only the input \textbf{natural-language problem} and final \textbf{solver outcome}, but also process-level supervision for the intermediate \textbf{modeling rationale} and reference \textbf{solver code}.
These intermediate elements provide direct guidance, yet they are costly to annotate and can tie the learned policy to the solver API and modeling patterns represented in the demonstrations.
Adapting SFT to a new solver typically requires reconstructing the \textbf{solver code} element and often the associated \textbf{modeling rationale} for another solver-specific reference dataset.

We instead seek to learn from only the \textbf{natural-language problem} and answer-level supervision derived from the \textbf{solver outcome}.
A solver can evaluate a generated program and expose whether the complete formulation produces the correct result, allowing RL to improve modeling without labeled \textbf{modeling rationales} or reference \textbf{solver code}.
This motivates \methodfull (\methodshort): the policy generates a formulation and solver program, receives feedback from the resulting solver outcome, and learns modeling strategies that can be reused across solver environments.

\subsection{Problem Formulation}
\label{sec:problem_formulation}
\label{sec:prompt}

Let $\mathcal{D}=\{(q,a)\}$ be a dataset of problem--answer pairs, where $q$ is a natural-language problem and $a$ is its ground-truth outcome.
The outcome is either a finite optimal objective $a\in\mathbb{R}$ or a symbolic status $a\notin\mathbb{R}$, such as \texttt{INFEASIBLE} or \texttt{UNBOUNDED}.
Given a target solver $s$, the task context is $x=(q,s)$, and a policy $\pi_\theta(y\mid x)$ generates a structured response.

\paragraph{Output protocol}
We use the prompt template in Figure~\ref{fig:prompt_template_clean}.
The solver used for the task is written into the \texttt{\{solver\}} placeholder, and the response contains a mathematical modeling rationale and executable Python code.
The program must print \texttt{Just print the best solution: \{solution value\}} for an optimal solution and \texttt{No Best Solution} otherwise.

\input{figures/main_prompt}

Each response follows a machine-parseable schema with exactly one \texttt{<think>} block followed by exactly one \texttt{<code>} block:
\begin{equation}
\label{eq:output_schema}
\begin{split}
y \equiv \
& \texttt{<think>}\ y_{\texttt{think}}\ \texttt{</think>} \\
& \texttt{<code>}\ y_{\texttt{code}}\ \texttt{</code>}.
\end{split}
\end{equation}
Thus, $y=(y_{\texttt{think}},y_{\texttt{code}})$, where $y_{\texttt{think}}$ is the modeling rationale and $y_{\texttt{code}}$ is the solver program.
A deterministic parser extracts only the unique code block; malformed responses are treated as non-executable.

\paragraph{Solver execution and observation}
The extracted program is evaluated in a solver environment $\mathcal{E}$:
\begin{equation}
o=\mathcal{E}(y_{\texttt{code}}), \qquad o=(c,\sigma,v,\ell),
\label{eq:obs_tuple}
\end{equation}
where $c\in\{0,1\}$ indicates execution success, $\sigma$ is the normalized solver status, $v\in\mathbb{R}$ is the reported objective value, and $\ell$ contains diagnostic logs.
The status determines the semantics of the observation: $v$ is meaningful when $\sigma=\text{\texttt{OPTIMAL}}$, whereas statuses such as \texttt{INFEASIBLE} and \texttt{UNBOUNDED} directly represent symbolic outcomes.
The resulting observation summarizes the end-to-end behavior of the generated model.
By comparing this observation with the labeled outcome, we obtain a reward signal for reinforcement learning.

\subsection{Reinforcement Learning from Solver Outcomes}
\label{sec:optimization}

We optimize the policy with RL by converting each generated response and its solver observation into a scalar reward.
The RL objective is
\begin{equation}
\label{eq:rl_objective}
\max_{\theta}\;
\mathbb{E}_{(q,a)\sim\mathcal{D}}
\ \mathbb{E}_{y\sim\pi_\theta(\cdot\mid x)}
\Big[\, R\big(y,\mathcal{E}(y_{\texttt{code}});a\big)\,\Big].
\end{equation}
Unlike process-supervised SFT, this objective never compares the generated rationale or code against annotated references.

\paragraph{Reward design}
The reward combines a lightweight format term with an outcome-correctness term:
\begin{equation}
\label{eq:reward_total}
R(y,o;a)= r_{\texttt{fmt}}(y) + r_{\texttt{ans}}(o; a).
\end{equation}
The format reward $r_{\texttt{fmt}}$ enforces the output protocol in Eq.~\eqref{eq:output_schema}, ensuring that model outputs are reliably parseable and executable.
We combine a soft tag-count check with a hard regular-expression match:
\begin{equation}
\label{eq:fmt_reward}
\begin{aligned}
r_{\texttt{fmt}}(y)=
\sum_{t\in\mathcal{T}} w_t\,\mathbb{I}[n_t(y)=1]
+ w_m M(y),
\end{aligned}
\end{equation}
where $\mathcal{T}=\{\texttt{<think>},\texttt{</think>},\texttt{<code>},\texttt{</code>}\}$, $n_t(y)$ counts occurrences of tag $t$ in $y$, $M(y)$ indicates whether $y$ matches the required block-ordering pattern, and $w_t,w_m\ge 0$ are fixed weights.
The soft tag-count term provides a dense learning signal during early training, while the hard regex constraint strictly enforces the required structure.
The answer reward is derived solely from the solver observation and labeled outcome.
Intuitively, $r_{\texttt{ans}}$ is a binary correctness signal:
\begin{equation}
\label{eq:ans_reward}
r_{\texttt{ans}}(o;a)=
\begin{cases}
1, & C(o,a),\\
0, & \text{otherwise}.
\end{cases}
\end{equation}
Here $C(o,a)$ means that the code executes successfully and the solver outcome agrees with the label.
For numeric labels, the solver must return an optimal value within tolerance; for symbolic labels, the normalized status must match the label.
All other cases are incorrect.

The numeric tolerance uses the smaller of the absolute error $|v-a|$ and the relative error $|v-a|/\max(|a|,\delta)$, and requires it to be below $\epsilon$.
We set $\epsilon=10^{-4}$ and $\delta=10^{-12}$ to avoid division by zero.
This sparse reward credits a response only when its complete generated model produces the correct solver outcome.

\input{algorithm}

Algorithm~\ref{alg:training} summarizes the resulting \textit{generate--execute--feedback--update} process.
For each problem, the policy samples a group of candidate outputs, obtains their solver observations and rewards, and converts the group-relative rewards into advantages.
We update the policy using GRPO~\cite{shao2024deepseekmathpushinglimitsmathematical}; importantly, the update uses only generated outputs, solver observations, and labeled outcomes.

\section{Cross-Solver Generalization}

Optimization modeling systems are rarely tied to a single solver in practice: different deployment settings may prefer different solvers because of licensing, performance, supported problem classes, or integration requirements.
This raises a question beyond learning to generate one solver's code: whether the learned modeling capability can be reused when the solver interface changes.
Since \methodshort{} trains from solver outcomes rather than solver-specific reference programs, the solver can be treated as a switchable execution environment.
The following sections describe how this environment abstraction is defined and how it enables several forms of cross-solver generalization.

\subsection{Switchable Solver Environments}
\label{sec:exec_framework}

Let $\mathcal{S}$ denote the set of supported solvers, such as \gurobi, \ortools, and \copt.
Different solvers expose different Python APIs, status names, and output conventions, but the learning objective only needs to know whether the generated model produces the labeled outcome.
We therefore separate solver-specific execution from solver-independent learning by associating each solver with an environment.
For each $s\in\mathcal{S}$, the environment
\begin{equation}
\mathcal{E}_s:\ y_{\texttt{code}}\mapsto o,
\qquad
o=\mathcal{E}_s(y_{\texttt{code}}).
\end{equation}
takes the generated solver code as input and returns the standardized observation $o$ defined in Eq.~\eqref{eq:obs_tuple}.
This environment absorbs solver-level differences: it invokes the selected solver interface, interprets solver-specific status codes, extracts the reported objective value when meaningful, and maps the result into the common tuple $(c,\sigma,v,\ell)$.
As a result, the reward in Eq.~\eqref{eq:reward_total} can be applied without solver-specific cases.

Switching solvers changes only two components: the solver name written in the prompt placeholder \texttt{\{solver\}} and the environment $\mathcal{E}_s$ used to evaluate the generated code.
The problem--answer data, output protocol, RL objective, and reward definition remain unchanged.
This separation lets \methodshort{} treat solver APIs as interchangeable environments while preserving a single outcome-supervised learning procedure.

\subsection{Cross-Solver Generalization Strategies}
\label{sec:generalization}

The switchable environment makes solver generalization a deployment choice rather than a change to the learning objective.
Depending on the target solver's API coverage and the available budget for additional training, \methodshort{} supports three complementary strategies.

\paragraph{Zero-shot training-free transfer}
We directly change the solver identifier and environment from a source solver $s_{\texttt{src}}$ to a target solver $s_{\texttt{tgt}}$ without parameter updates.
This is the lowest-cost setting: it uses the same problem--answer data, policy parameters, output protocol, and reward definition, and only changes the solver name written in the prompt and the environment used for execution.
It is most suitable when the target solver has sufficient overlap with the source solver in the model's pretraining distribution or when a quick deployment test is needed.
Because the policy has never been trained in the target environment, this setting directly tests whether learned optimization modeling competence transfers beyond source-solver syntax.

\paragraph{Direct RL training}
When target-environment execution is available, we train directly in that environment using solver-outcome rewards, without constructing target-solver reference programs.
This setting still requires only problem--answer pairs $(q,a)$: the generated code is executed by $\mathcal{E}_{s_{\texttt{tgt}}}$, and the same answer reward evaluates whether the resulting solver observation matches $a$.
Thus, direct RL can specialize the policy to target-solver syntax and runtime behavior while preserving outcome-only supervision.
More generally, RL training over one or more target environments is
\begin{equation}
\label{eq:target_environment_objective}
\max_{\theta}\;
\mathbb{E}_{(q,a)\sim\mathcal{D}}
\ \mathbb{E}_{s\sim p_{\texttt{tgt}}(s)}
\ \mathbb{E}_{y\sim\pi_\theta(\cdot\mid x)}
\Big[\, R\big(y, \mathcal{E}_s(y_{\texttt{code}}); a\big)\,\Big],
\end{equation}
where $p_{\texttt{tgt}}(s)$ is the solver sampling distribution over the target environment.
For single-environment training, $p_{\texttt{tgt}}$ always selects the target solver $s_{\texttt{tgt}}$; for multi-environment training, it can mix several target environments so that one policy is optimized against multiple solver interfaces.

\paragraph{SFT-cold-started RL}
For under-covered solvers with scarce pretraining exposure and near-zero initial generation success, direct RL may suffer from a cold-start problem: the model rarely produces executable code, resulting in sparse positive feedback and slow convergence.
In this case, we use a small SFT seed only to cross the API barrier, not as the main supervision source.
We first use a general-purpose LLM and target-solver documentation to translate a small set of source-solver programs:
\begin{equation}
    (q, s_{\texttt{src}}, y_{\texttt{src}}) \ \xrightarrow{\texttt{LLM Translation}} \ (q, s_{\texttt{tgt}}, y_{\texttt{tgt}}).
\end{equation}
Only translations whose outcomes match the ground truth under $\mathcal{E}_{s_{\texttt{tgt}}}$ are retained.
A brief SFT stage teaches basic target-solver API usage and output conventions, after which RL continues in the target environment with the same solver-outcome reward.
This strategy reduces exploration difficulty for under-covered solvers while keeping solver outcomes, rather than translated code, as the main source of optimization pressure.

\section{Experiments}

\subsection{Experimental Settings}
\label{sec:exp_settings}

\paragraph{Datasets}

We construct the training set $\mathcal{D}$ based on OR-Instruct-3K from ORLM~\cite{huang2025orlm}, which contains 3,000 optimization problems.
To align with our \textit{outcome-only} formulation, we retain only the natural-language descriptions $q$ and extract the ground-truth outcomes $a$ (optimal objective values) by executing the original reference programs.
All intermediate process-level annotations (e.g., variable derivations and reference code) are discarded during training.
We evaluate \methodshort\ across four representative benchmarks: NL4OPT~\cite{ramamonjison2022nl4opt}, MAMO~\cite{huang2025mamo}, IndustryOR~\cite{huang2025orlm}, and OptiBench~\cite{yang2025optibench}. 
Notably, for the MAMO ComplexLP subset, we identified and corrected erroneous reference answers in several Traveling Salesman Problem instances where the original programs provided wrong results.

\paragraph{Baseline Methods}
We compare \methodshort\ against two distinct paradigms.
Prompting-based models include various strategies applied to GPT-4o~\cite{hurst2024gpt4o}, including Standard Prompting, Chain-of-Thought (CoT)~\cite{wei2022chain}, Chain-of-Experts (CoE)~\cite{xiao2024chainofexperts}, and the agentic system OptiMUS~\cite{Ahmaditeshnizi2024optimus}.
We compare against ORLM~\cite{huang2025orlm}, which represents the \textit{process-level supervision} paradigm via supervised fine-tuning (SFT) on the full OR-Instruct dataset.

\paragraph{Evaluation Metric}
We report Accuracy, defined as the fraction of test instances where the predicted outcome $v$ from the execution observation $o$ matches the ground-truth $a$. 
Following previous work~\cite{huang2025orlm}, a numeric prediction $v$ is counted as correct if either its absolute error or relative error is within the tolerance $\epsilon_{\text{eval}}$:
\begin{equation}
\label{eq:eval_metric}
\bigl(|v-a| \le \epsilon_{\text{eval}}\bigr)
\ \lor\
\biggl(\frac{|v-a|}{\max(|a|,\delta)} \le \epsilon_{\text{eval}}\biggr).
\end{equation}
We set $\epsilon_{\text{eval}}=0.05$ for all results in the main paper.

\paragraph{Implementation Details}
Table~\ref{tab:impl_details} summarizes the default training and reward settings.
We organize the remaining implementation details along three components.

\textit{Solver environments:}
we instantiate \gurobi~\cite{gurobi}, \ortools~\cite{ortools}, and \copt~\cite{copt};
the main experiments use \gurobi\ and \ortools, while \copt\ is evaluated with an SFT cold start in Section~\ref{sec:exp_cold_start_copt}.
Each environment is a sandboxed Python execution harness with a 10-second timeout and 2-GB memory limit.
It exposes the selected solver through its standard Python interface, captures process outputs and solver logs, and normalizes them into $o=(c,\sigma,v,\ell)$.

\textit{Training setup:}
training-based models use Qwen2.5-7B~\cite{qwen2025qwen25technicalreport} as the base policy $\pi_\theta$, and ORLM~\cite{huang2025orlm} is reproduced with the same backbone following its original SFT protocol.
RL training is conducted on 4 NVIDIA A800 GPUs using verl\footnote{https://github.com/verl-project/verl}.

\textit{Generation and evaluation:}
RL rollout sampling uses temperature $=1$ and top-$p=1$; all evaluations use greedy decoding with temperature $=0$ and the accuracy criterion in Eq.~\eqref{eq:eval_metric}.

\begin{table}[t]
\centering
\caption{Default implementation details for training-based experiments.}
\label{tab:impl_details}
\small
\setlength{\tabcolsep}{4pt}
\begin{tabularx}{\columnwidth}{@{}lX@{}}
\toprule
\textbf{Item} & \textbf{Setting} \\
\midrule
Learning rate & $1\times10^{-6}$ \\
Batch size / group size & 64 / $G=8$ \\
RL rollout sampling & temperature $=1$, top-$p$ $=1$ \\
GRPO clipping & $\epsilon_{\text{clip}}=0.2$ \\
KL coefficient & $\beta=0.001$ \\
Training length & 1 epoch \\
Maximum sequence length & 8,192 tokens \\
Format reward weights & $w_t=0.125$ for each tag, $w_m=0.5$ \\
\bottomrule
\end{tabularx}
\end{table}

\subsection{Main Results}

\input{tables/main_result}

Table~\ref{tab:main_result} shows that solver-outcome RL enables \methodshort to perform even slightly better than process-supervised training on average, despite using no process-level annotations. 
Using the same Qwen2.5-7B backbone, \methodshort improves the base model by 18.15 average points and slightly surpasses the process-supervised ORLM baseline on average (60.48 vs. 59.38). 
It outperforms ORLM on OptiBench, MAMO ComplexLP, and IndustryOR, while achieving comparable performance on NL4OPT and MAMO EasyLP.

This comparison is particularly meaningful because ORLM relies on process annotations, including intermediate modeling rationales and reference solver programs, whereas \methodshort only uses final answers together with solver outcomes for reward computation. 
The fact that outcome-only RL is even slightly better on average suggests that solver outcomes provide an informative supervision signal for learning variables, objectives, and constraints at scale. 
More importantly, this design reduces the dependence on solver-specific expert traces, which are costly to collect and difficult to maintain when benchmarks or solver interfaces change.

\subsection{Ablation on Reasoning Traces}
\label{sec:exp_think_ablation}

\input{tables/think_ablation}

This ablation investigates the effect of explicit reasoning in outcome-supervised learning.
Specifically, we examine whether outcome supervision suffices for a direct problem-to-code mapping, or if an explicit reasoning step is essential for successful alignment.
To this end, we compare the standard \methodshort\ against a reasoning-free variant that bypasses the \texttt{<think>} block and generates code directly.

As shown in Table~\ref{tab:think_ablation}, removing the reasoning block triggers a significant performance collapse across most benchmarks.
This degradation is particularly acute on logically demanding tasks such as IndustryOR and MAMO ComplexLP, indicating that a direct mapping from problem descriptions to code often fails to capture complex optimization logic. In contrast, on the simpler MAMO EasyLP subset, the absence of reasoning yields a marginal performance gain, suggesting that explicit reasoning might introduce unnecessary overhead or ``overthinking'' for trivial instances.
These results affirm that in outcome-supervised optimization modeling, the \texttt{<think>} block acts not just as a post-hoc explanation, but as an essential internal workspace:
outcome-level rewards optimize code generation indirectly by reinforcing the intermediate logical steps that are necessary to produce a correct solver outcome.

\subsection{Cross-Solver Generalization}
\label{sec:exp_cross_solver}

We evaluate the three cross-solver strategies introduced in Section~\ref{sec:generalization}: zero-shot training-free transfer, direct RL training, and SFT-cold-started RL for an under-covered solver.

\subsubsection{Zero-Shot Training-Free Solver Transfer}
\label{sec:exp_zeroshot}

To examine whether \methodshort\ elicits universal mathematical modeling principles, we evaluate a policy trained exclusively on \gurobi ($s_{\texttt{src}}$) by directly switching the execution backend to \ortools\ ($s_{\texttt{tgt}}$) without any parameter updates. This setting tests the model's ability to decouple underlying mathematical logic from solver-specific syntactic requirements.

\input{tables/zerosolver}

As shown in Table~\ref{tab:zero}, \methodshort demonstrates superior zero-shot training-free transfer compared to the SFT-based baseline.
While the process-supervised ORLM suffers from a near-complete collapse on the target solver (e.g., dropping to 0.00\% accuracy on MAMO EasyLP), \methodshort\ retains usable performance and consistently outperforms the target-solver base model across all benchmarks.
This stark contrast suggests that process-level supervision often forces the model to mimic specific solver APIs and code patterns present in the training traces, leading to brittle, syntax-dependent policies. In contrast, by optimizing solver-outcome rewards, \methodshort\ learns to prioritize invariant mathematical structures. This allows the model to function as a reusable optimization modeler across solver interfaces without specialized fine-tuning for every new environment.

\subsubsection{Direct RL Training Across Solver Environments}
\label{sec:exp_direct_rl}

\input{tables/different_solver}

We next evaluate direct RL training in diverse solver environments.
Starting from the same base model, we run the identical \textit{generate--execute--feedback--update} loop in each solver environment $\mathcal{E}_s$.
Concretely, switching the solver amounts to changing only the execution backend and the induced observation $o=(c,\sigma,v,\ell)$, while keeping the training data as outcome-only pairs $(q,a)$ and \emph{without} constructing solver-specific reference programs.

Table~\ref{tab:different_solver} compares performance before and after solver-feedback RL in each solver environment.
For \gurobi, \methodshort\ improves all benchmarks.
For \ortools, it improves four out of five benchmarks, with particularly clear gains on MAMO EasyLP, NL4OPT, and IndustryOR, while MAMO ComplexLP decreases slightly.
These results indicate that deterministic solver feedback provides a strong signal for training the code generator in a target environment, but the benefit can vary by benchmark and solver interface.
These results support our central claim:
training for a new solver requires only switching the environment, rather than rebuilding solver-specific supervised datasets.

\subsubsection{SFT-Cold-Started RL for an Under-covered Solver}
\label{sec:exp_cold_start_copt}

\input{tables/copt_cold_start}

We next evaluate the cold-start strategy on \copt, a commercial solver with limited coverage in LLM pretraining corpora.
We compare three settings: (i) the base model with zero-shot prompting, (ii) the base model fine-tuned with a small cold-start SFT set, and (iii) the cold-started model further improved through RL.
Following the SFT-cold-started RL strategy in Section~\ref{sec:generalization}, the cold-start SFT set contains only 100 randomly sampled problem-code pairs, constructed by translating \gurobi reference programs into \copt programs and filtering them by execution correctness.

Table~\ref{tab:copt_cold_start} shows that this small SFT cold start substantially improves over the zero-shot baseline across all benchmarks, confirming that minimal syntactic supervision can overcome the initial API barrier for an under-covered solver.
More importantly, subsequent RL further improves every benchmark over SFT alone, demonstrating that solver feedback continues to refine modeling correctness beyond static API imitation.
These results show that SFT-cold-started RL provides a practical route to modeling with under-covered solvers: a small SFT seed establishes API priors, after which solver-outcome RL improves modeling correctness.

\subsection{Analysis of Training Dynamics}
\label{sec:training_dynamics}
\input{figures/dynamic}

We further inspect the training trajectory to understand how solver-outcome RL changes the model's generation behavior over time.
In particular, we track response lengths, task accuracy, and reward components during rollout training.

Figure~\ref{fig:dynamic}(a) shows that \textit{thinking} length increases as accuracy improves, while \textit{code} length remains broadly stable and fluctuates around a similar level.
Thus, RL primarily expands the model's explicit modeling deliberation without systematically increasing or decreasing implementation length.
Because no reasoning traces or reference programs are provided during training, the longer \texttt{<think>} block is not produced by imitating annotated processes; it emerges from optimizing solver outcomes.
The stable code length further suggests that the accuracy gains cannot be explained simply by generating longer or shorter programs.

Figure~\ref{fig:dynamic}(b) reveals a complementary two-stage learning pattern.
The format rewards rise and saturate early, showing that the policy quickly learns the machine-parseable interface required for execution.
The correctness reward continues to improve after format compliance has largely stabilized, indicating that later RL updates primarily improve the generated optimization models rather than merely repairing output structure.
Together, these dynamics show how the format reward first establishes a reliable interaction channel, after which the solver-outcome reward drives the harder improvement in modeling correctness.
The mismatch between increasing reasoning length and stable code length is also informative: the policy appears to allocate more tokens to interpreting constraints and objectives, rather than simply producing larger programs.
This pattern is consistent with the reward design, since only executable models with correct outcomes receive answer credit.
Output length alone does not establish reasoning quality, but its co-evolution with accuracy and the human analysis in Section~\ref{sec:human_analysis} provide consistent evidence that RL changes the modeling process beyond surface formatting.

\subsection{Human Analysis of Reasoning and Errors}
\label{sec:human_analysis}

\input{tables/human}

To better understand what solver-feedback RL improves, we conduct a fine-grained human analysis on 516 OptiBench samples.
Two mathematics graduate students independently annotate reasoning and implementation quality, with disagreements resolved by a senior researcher; the inter-annotator agreement is high.
We focus on two core modeling components: constraints and objective functions.
For constraints, annotators evaluate both reasoning correctness and code implementation, including whether key implicit restrictions are missed.
For objective functions, annotators evaluate whether the mathematical objective and its code implementation match the problem intent.

Table~\ref{tab:human} shows that \methodshort{} improves reasoning quality rather than merely optimizing executable syntax.
Compared with the base model, \methodshort{} improves constraint reasoning from 24.6\% to 26.4\% and objective reasoning from 74.2\% to 77.1\%.
It also achieves the best objective-code accuracy, increasing from 81.4\% to 87.8\%.
ORLM obtains stronger constraint-code accuracy, which is expected because it is trained with reference programs, but its reasoning scores are lower than or comparable to the base model.
This contrast suggests that process supervision can encourage code-pattern imitation, whereas outcome-supervised RL better preserves and improves independent modeling reasoning.

\input{tables/error_analysis}

Table~\ref{tab:error_analysis} further breaks down the error distribution before and after RL training.
The fully correct rate increases from 51.55\% to 63.37\%, while implementation errors and variable errors drop substantially.
The largest reduction is in variable errors, from 6.59\% to 0.78\%, indicating that solver feedback is especially effective for mistakes that directly affect feasibility and objective values, such as missing integrality or incorrect bounds.
Constraint and objective errors decrease more modestly, showing that some residual failures require deeper semantic understanding beyond what scalar solver feedback can fully diagnose.
Overall, the human analysis indicates that \methodshort{} improves both executable correctness and modeling quality, with the strongest gains on errors that are directly exposed by solver outcomes.

\section{Conclusion}

We studied how RL from solver outcomes can improve optimization modeling and support generalization across solver environments.
\methodfull{} (\methodshort) uses only problem--answer pairs and solver outcomes, eliminating the need for annotated reasoning traces and reference programs.
With the same Qwen2.5-7B backbone, it slightly surpasses the process-supervised ORLM baseline on average while producing solver-executable programs.
We further demonstrate zero-shot training-free transfer, direct RL training in multiple solver environments, and SFT-cold-started RL for an under-covered solver without reconstructing full solver-specific reference datasets.
Direct RL provides strong but not uniform gains across benchmarks, reflecting remaining differences between solver interfaces and modeling tasks.
Training dynamics, reasoning ablations, and human analysis suggest that the improvements extend beyond format and API compliance to the underlying modeling process, although deeper constraint and objective errors remain challenging.
Overall, solver outcomes provide a practical RL signal for reducing process supervision and extending optimization modeling across heterogeneous solver environments.

\section*{Acknowledgments}
This work was supported by the
National Science and Technology Major Project under Grant No. 2026ZD1614300, and by the National Natural Science Foundation
of China under Grant Nos. 62476134 and 62606222.

\bibliographystyle{IEEEtran}
\bibliography{mybib}

\end{document}

%% file: figures/framework.tex
\begin{figure*}[t]
    \centering
    \includegraphics[width=0.9\textwidth]{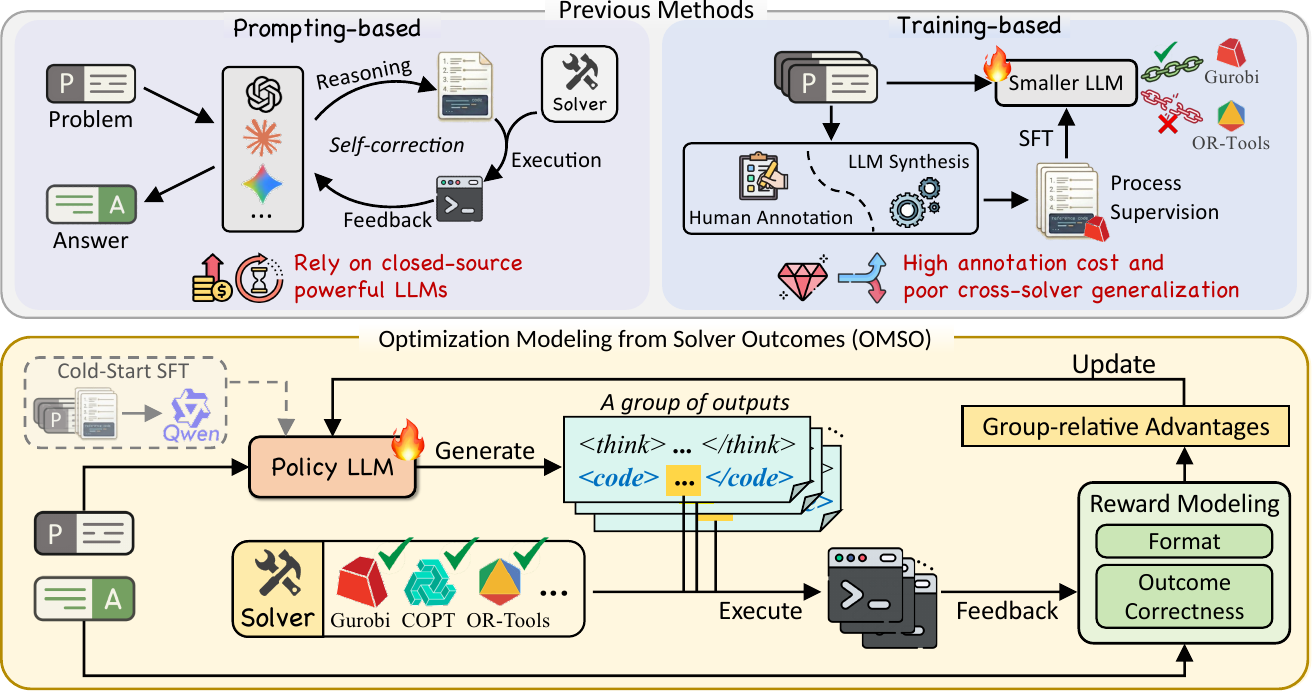}
    \caption{Optimization modeling under different supervision and solver-transfer settings.}
    \label{fig:framework}
\end{figure*}

%% file: figures/preliminary_case2.tex
\begin{figure}[t]
\centering
\begin{tcolorbox}[
    enhanced,
    colback=white,
    colframe=black!50,
    boxrule=0.6pt,
    arc=2pt,
    width=\columnwidth,
    title={\textbf{Task Example:} From Natural Language to a Solver Outcome},
    coltitle=black,
    colbacktitle=gray!15,
    fonttitle=\small\bfseries,
    left=5pt, right=5pt, top=5pt, bottom=5pt
]
\scriptsize
\textcolor{blue!70!black}{\textbf{Natural-language problem.}}
A golf course can transport guests using golf carts and pull carts. A golf cart carries 4 guests, while a pull cart carries 1 guest. At most 60\% of all carts can be golf carts. The goal is to transport at least 80 guests with the minimum total number of carts.

\tcbline

\textcolor{orange!70!black}{\textbf{Modeling rationale.}}

Let $x$ be the number of golf carts and $y$ be the number of pull carts.

\textbf{Objective:} Minimize $x + y$

\textbf{Constraints:}
\begin{itemize}[nosep, leftmargin=*]
    \item Guest capacity: $4x + y \geq 80$
    \item Golf cart ratio: $x \leq 0.6(x + y) \Rightarrow y \geq \frac{2}{3}x$
    \item Non-negativity: $x, y \geq 0$ and integer
\end{itemize}

\tcbline

\textcolor{purple!70!black}{\textbf{Solver code.}}
\begin{lstlisting}[style=pythonstyle, basicstyle=\ttfamily\tiny, breaklines=true]
import gurobipy as gp
from gurobipy import GRB

m = gp.Model("golf_carts")
x = m.addVar(name="golf_carts", vtype=GRB.INTEGER, lb=0)
y = m.addVar(name="pull_carts", vtype=GRB.INTEGER, lb=0)

m.setObjective(x + y, GRB.MINIMIZE)
m.addConstr(4*x + y >= 80)
m.addConstr(y >= (2/3)*x)
m.optimize()

if m.status == GRB.OPTIMAL:
    print(f"Just print the best solution: {m.objVal}")
\end{lstlisting}

\tcbline

\textcolor{green!60!black}{\textbf{Solver outcome.}}
Running the code returns \texttt{OPTIMAL} with objective value $29.0$.
\end{tcolorbox}
\caption{An example of optimization modeling, from interpreting a natural-language problem to constructing and running a solver program.}
\label{fig:preliminary_case2}
\end{figure}

%% file: figures/main_prompt.tex
\begin{figure}[t]
    \centering
    \begin{minipage}{0.99\columnwidth}
        \centering
        \begin{tcolorbox}[
            enhanced,
            colback=black!3,
            colframe=black!40,
            boxrule=0.5pt,
            arc=0pt,
            sharp corners,
            left=10pt, right=10pt, top=10pt, bottom=10pt,
            fontupper=\small
        ]
            Below is an Operations Research question. Build a mathematical model and corresponding Python code using \texttt{\{solver\}} to address the question.
            \vspace{0.5em}

            In the Python code, please include a final logic block as follows:
            if the problem has an optimal solution, print ``\texttt{Just print the best solution: \{solution value\}}'';
            otherwise, print ``\texttt{No Best Solution}''.
            \vspace{0.5em}

            The response consists of two parts. The reasoning process (including the mathematical model) and the Python code must be enclosed within \texttt{<think>}...\texttt{</think>} and \texttt{<code>}...\texttt{</code>} tags, respectively:
            \vspace{0.5em}

            \texttt{<think>}
            \newline
            \textit{reasoning process here, including mathematical model...}
            \newline
            \texttt{</think>}
            \vspace{0.5em}

            \texttt{<code>}
            \newline
            \textit{your response here, including Python code...}
            \newline
            \texttt{</code>}
            \vspace{0.5em}

            Here is the question:
            \newline
            \texttt{\{Question\}}
        \end{tcolorbox}
    \end{minipage}
    \caption{Prompt template with a solver placeholder and a fixed two-block output format for parsing and execution.}
    \label{fig:prompt_template_clean}
\end{figure}

%% file: algorithm.tex
\begin{algorithm}[ht]
\small
\renewcommand{\algorithmiccomment}[1]{\hfill\textit{// #1}}
\caption{\methodfull{} Training via Generate--Execute--Feedback--Update}
\label{alg:training}
\begin{algorithmic}[1]
\REQUIRE Dataset $\mathcal{D}=\{(q,a)\}$; target solver $s$; policy $\pi_\theta$;
         batch size $B$; group size $G$; solver environment $\mathcal{E}(\cdot)$; rewards $r_{\texttt{fmt}}(\cdot)$ (Eq.~(\ref{eq:fmt_reward})), $r_{\texttt{ans}}(\cdot;\cdot)$ (Eq.~(\ref{eq:ans_reward})); GRPO optimization settings.

\WHILE{not converged}
  \STATE Sample $\{(q_j,a_j)\}_{j=1}^B \sim \mathcal{D}$.
  \STATE Freeze $\pi_{\theta_{\texttt{old}}} \leftarrow \pi_\theta$.
  \FOR{$j=1$ {\bfseries to} $B$}
    \STATE $x_j \leftarrow (q_j,s)$.
    \STATE Sample $\{y_{j,k}\}_{k=1}^G \sim \pi_{\theta_{\texttt{old}}}(\cdot \mid x_j)$.
    \FOR{$k=1$ {\bfseries to} $G$}
      \STATE $r^{\texttt{fmt}}_{j,k} \leftarrow r_{\texttt{fmt}}(y_{j,k})$.
      \STATE Parse the \emph{unique} \texttt{<code>} block from $y_{j,k}$ (Sec.~\ref{sec:prompt}).
      \IF{parsing fails (schema violation)}
        \STATE $o_{j,k} \leftarrow (0,\texttt{ERROR},0,\ell)$ \hfill // non-executable
      \ELSE
        \STATE Execute: $o_{j,k} \leftarrow \mathcal{E}(y^{\texttt{code}}_{j,k})$.
      \ENDIF
      \STATE $r^{\texttt{ans}}_{j,k} \leftarrow r_{\texttt{ans}}(o_{j,k};a_j)$.
      \STATE $R_{j,k} \leftarrow r^{\texttt{fmt}}_{j,k} + r^{\texttt{ans}}_{j,k}$ \hfill (Eq.~(\ref{eq:reward_total}))
    \ENDFOR

    \STATE Compute group stats $\mu_R,\sigma_R$ over $\{R_{j,k}\}_{k=1}^G$.
    \STATE $A_{j,k} \leftarrow \dfrac{R_{j,k}-\mu_R}{\sigma_R+\delta_A}$ for $k=1..G$.
  \ENDFOR

  \STATE Update $\theta$ with GRPO using the sampled contexts $x$, outputs $y$, and group-relative advantages $A$.
\ENDWHILE
\end{algorithmic}
\end{algorithm}

%% file: tables/main_result.tex
\begin{table*}[t]
  \centering
  \small
  \caption{Performance comparison on optimization modeling benchmarks using \gurobi. ORLM uses process annotations, whereas \methodshort uses no process annotations. ORLM denotes our reproduced result under the same evaluation setup.}
  \label{tab:main_result}
  \begin{tabular}{l *{6}{S[table-format=2.2]}}
    \toprule
    \textbf{Model} & {\textbf{OptiBench}} & {\textbf{NL4OPT}} & {\textbf{MAMO EasyLP}} & {\textbf{MAMO ComplexLP}} & {\textbf{IndustryOR}} & {\textbf{Avg.}} \\
    \midrule
    \multicolumn{7}{l}{\textit{\textbf{Prompting-based Method}}} \\
    \addlinespace[2pt]
    GPT-4o (Standard)& 42.30 & 70.30 & 84.30 & 58.80 & 27.00 & 56.54 \\
    GPT-4o (CoT)     & 42.00 & 71.60 & 84.80 & 59.90 & 29.00 & 57.46 \\
    GPT-4o (CoE)     & 43.20 & 76.40 & 85.70 & 64.00 & 34.00 & 60.66 \\
    OptiMUS          & 45.80 & 82.00 & 85.10 & 64.90 & 34.00 & 62.36 \\
    Qwen2.5-7B (Base)        & 41.19 & 55.10 & 65.95 & 28.43 & 21.00 & 42.33 \\
    \midrule 
    \multicolumn{7}{l}{\textit{\textbf{SFT with Process Annotations (Qwen2.5-7B)}}} \\
    \addlinespace[2pt]
    ORLM       & 60.96 & \bfseries 84.89 & \bfseries 88.34 & 35.71 & 27.00 & 59.38 \\
    \midrule 
    \multicolumn{7}{l}{\textit{\textbf{Outcome-Supervised RL without Process Annotations (Qwen2.5-7B)}}} \\
    \rowcolor{black!3} 
    \textbf{\methodshort} & \bfseries 62.95 & 84.08 & 88.19 & \bfseries 36.19 & \bfseries 31.00 & \bfseries 60.48 \\
    \bottomrule
  \end{tabular}
\end{table*}

%% file: tables/think_ablation.tex
\begin{table}[t]
    \centering
    \caption{Effect of explicit reasoning traces on the base model and \methodshort. The comparison tests whether solver-outcome RL benefits from an explicit modeling rationale before code generation.}
    \label{tab:think_ablation}
    \small
    \setlength{\tabcolsep}{2pt}
    \begin{tabular}{l *{5}{S[table-format=2.2]}}
        \toprule
        \textbf{Model} & {\textbf{Opti.}} & {\textbf{NL4OPT}} & {\textbf{MAMO-E}} & {\textbf{MAMO-C}} & {\textbf{Ind.OR}} \\
        \midrule
        Base w/o think & 46.67 & 59.18 & 83.89 & 16.66 & 15.00 \\
        Base w/ think & 59.30 & 80.00 & 83.43 & 33.17 & 27.00 \\
        \midrule 
        \methodshort{} w/o think & 53.74 & 76.32 & \bfseries 90.03 & 20.37 & 18.00 \\
        \methodshort{} w/ think & \bfseries 62.95 & \bfseries 84.08 & 88.19 & \bfseries 36.19 & \bfseries 31.00 \\
        \bottomrule
    \end{tabular}
\end{table}

%% file: tables/zerosolver.tex
\begin{table}[t]
    \centering
    \small
    \setlength{\tabcolsep}{3pt}
    \caption{Zero-shot training-free transfer from \gurobi to the unseen \ortools environment. All models are evaluated on \ortools without target-environment parameter updates.}
    \label{tab:zero}
    \begin{tabular}{l *{5}{S[table-format=2.2]}}
        \toprule
        \textbf{Model} &  {\textbf{Opti.}} & {\textbf{NL4OPT}} & {\textbf{MAMO-E}} & {\textbf{MAMO-C}} & {\textbf{Ind.OR}} \\
        \midrule
        Base  & 40.86 & 59.18 & 53.83 & 18.48 & 20.00 \\
        
        ORLM  & 3.49 & 4.89 & 0.00 & 1.42 & 6.00 \\
        \methodshort & \bfseries 54.31 & \bfseries 77.55 & \bfseries 84.81 & \bfseries 22.27 & \bfseries 24.00 \\
        \bottomrule
    \end{tabular}
\end{table}

%% file: tables/different_solver.tex
\begin{table}[t]
    \centering
    \small
    \setlength{\tabcolsep}{1.5pt}
    \caption{Direct RL training in \gurobi and \ortools environments. Each policy is trained from the same base model Qwen2.5-7B using only outcomes from its designated solver environment.}
    \label{tab:different_solver}
    \begin{tabular}{l *{5}{S[table-format=2.2]}}
        \toprule
        \textbf{Solver / Model} & {\textbf{Opti.}} & {\textbf{NL4OPT}} & {\textbf{MAMO-E}} & {\textbf{MAMO-C}} & {\textbf{Ind.OR}} \\
        \midrule
        \textbf{\gurobi} & & & & & \\
        \quad Base & 41.19 & 55.10 & 65.95 & 28.43 & 21.00 \\
        \quad \methodshort  & \bfseries 62.95 & \bfseries 84.08 & \bfseries 88.19 & \bfseries 36.19 & \bfseries 31.00 \\
        \midrule
        \textbf{\ortools} & & & & & \\
        \quad Base & 40.86 & 59.18 & 53.83 & \bfseries 18.48 & 20.00 \\
        \quad \methodshort  & \bfseries 52.49 & \bfseries 77.14 & \bfseries 82.82 & 16.11 & \bfseries 31.00 \\
        \bottomrule
    \end{tabular}
\end{table}

%% file: tables/copt_cold_start.tex
\begin{table}[t]
    \centering
    \caption{SFT-cold-started RL for the under-covered \copt environment. A small SFT seed establishes API knowledge before solver-outcome RL.}
    \label{tab:copt_cold_start}
    \small
    \setlength{\tabcolsep}{2pt}
    \begin{tabular}{l *{5}{S[table-format=2.2]}}
        \toprule
        \textbf{Model} & {\textbf{Opti.}} & {\textbf{NL4OPT}} & {\textbf{MAMO-E}} & {\textbf{MAMO-C}} & {\textbf{Ind.OR}} \\
        \midrule
        Base & 0.00 & 0.00 & 0.00 & 0.00 & 0.00 \\
        SFT only & 57.30 & 81.22 & 87.11 & 34.12 & 24.00 \\
        SFT + RL& \bfseries 62.79 & \bfseries 85.30 & \bfseries 87.58 & \bfseries 39.33 & \bfseries 29.00 \\
        \bottomrule
    \end{tabular}
\end{table}

%% file: figures/dynamic.tex
\begin{figure}[t]
  \centering

  \begin{subfigure}{\columnwidth}
    \centering
    \includegraphics[width=\linewidth]{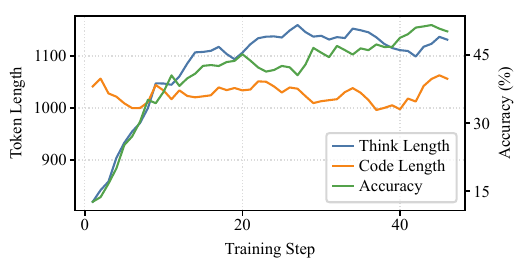}
    \caption{Thinking length increases while code length remains broadly stable as accuracy improves.}    \label{fig:dynamic:length}
  \end{subfigure}
  \begin{subfigure}{\columnwidth}
    \centering
    \includegraphics[width=\linewidth]{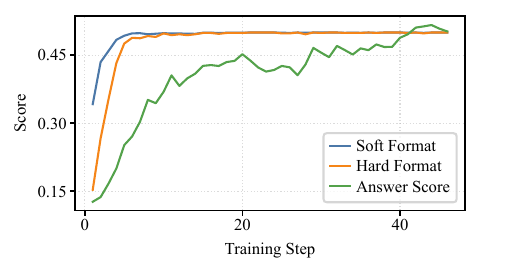}
    \caption{Format reward saturates early, whereas correctness reward rises gradually.}
    \label{fig:dynamic:scores}
  \end{subfigure}

  \caption{Training dynamics of solver-feedback RL.}
  \label{fig:dynamic}
\end{figure}

%% file: tables/human.tex
\begin{table}[ht]
    \centering
    \small
    \caption{Human evaluation accuracy on reasoning quality and code implementation.}
    \label{tab:human}
    \setlength{\tabcolsep}{3pt}
    \renewcommand{\arraystretch}{1.15}
    \begin{tabular}{l *{4}{S[table-format=2.1]}}
        \toprule
        \multirow{2}{*}{\textbf{Model}} &
        \multicolumn{2}{c}{\textbf{Constraints}} &
        \multicolumn{2}{c}{\textbf{Objective Function}} \\
        \cmidrule(lr){2-3}\cmidrule(lr){4-5}
        & {\textbf{Reasoning}} & {\textbf{Code}}
        & {\textbf{Reasoning}} & {\textbf{Code}} \\
        \midrule
        Qwen2.5-7B     & 24.6 & 58.3 & 74.2 & 81.4 \\
        ORLM    & 23.4 & \bfseries 78.7 & 65.4 & 78.7 \\
        \methodshort & \bfseries 26.4 & 59.7 & \bfseries 77.1 & \bfseries 87.8 \\
        \bottomrule
    \end{tabular}
\end{table}

%% file: tables/error_analysis.tex
\begin{table}[t]
    \centering
    \small
    \caption{Fine-grained error analysis on 516 sampled instances from OptiBench.}
    \label{tab:error_analysis}
    \setlength{\tabcolsep}{3pt}
    \renewcommand{\arraystretch}{1.12}
        \begin{tabular}{l S[table-format=2.2] S[table-format=2.2] S[table-format=+2.2]}
            \toprule
            \textbf{Error Type} & {\textbf{Before RL (\%)}} & {\textbf{After RL (\%)}} & {\textbf{Diff (\%)}} \\
            \midrule
            Implementation & 10.85 & 7.17  & -3.68 \\
            Objective      & 12.98 & 12.60 & -0.38 \\
            Constraint     & 18.02 & 16.09 & -1.93 \\
            Variable       & 6.59  & 0.78  & -5.81 \\
            Fully Correct  & 51.55 & 63.37 & +11.82 \\
            \bottomrule
        \end{tabular}
\end{table}